\documentclass[a4paper,twoside]{article}

\usepackage[margin=2cm]{geometry}
\usepackage{adjustbox}
\usepackage{multirow}
\usepackage[inline]{enumitem}

\usepackage{hyperref} 
\usepackage[utf8]{inputenc}
\usepackage{dblfloatfix}
\usepackage{epsfig}
\usepackage{subcaption}
\usepackage{calc}
\usepackage{amssymb}
\usepackage{amstext}
\usepackage{amsmath}
\usepackage{amsthm}
\usepackage{multicol}
\usepackage{pslatex}
\usepackage{apalike}
\usepackage{SCITEPRESS}     

\begin{document}

\title{Intelligent Roundabout Insertion using Deep Reinforcement Learning}

\author{Anonymous Author(s)}
\author{\authorname{Alessandro Paolo Capasso\sup{1}, Giulio Bacchiani\sup{1} and Daniele Molinari\sup{2}}
\affiliation{\sup{1}Vislab - University of Parma, Parma, Italy}
\affiliation{\sup{2}Vislab, Parma, Italy}
\email {alessandro.capasso@unipr.it, giulio.bacchiani@studenti.unipr.it, dmolina@ce.unipr.it}
}

\keywords{Autonomous Driving, Deep Reinforcement Learning, Multi-Agent Systems, Agent Cooperation and Negotiation, Maneuver Planning System.}

\abstract{An important topic in the autonomous driving research is the development of maneuver planning systems. Vehicles have to interact and negotiate with each other so that optimal choices, in terms of time and safety, are taken. For this purpose, we present a maneuver planning module able to negotiate the entering in busy roundabouts. The proposed module is based on a neural network trained to predict when and how entering the roundabout throughout the whole duration of the maneuver.
Our model is trained with a novel implementation of \textit{A3C}, which we will call \textit{Delayed A3C} (D-A3C), in a synthetic environment where vehicles move in a realistic manner with interaction capabilities.
In addition, the system is trained such that agents feature a unique tunable behavior, emulating real world scenarios where drivers have their own driving styles. Similarly, the maneuver can be performed using different aggressiveness levels, which is particularly useful to manage busy scenarios where conservative rule-based policies would result in undefined waits.}

\onecolumn \maketitle \normalsize \setcounter{footnote}{0} \vfill

\section{\uppercase{Introduction}}
\label{sec:introduction}
\noindent The study and development of autonomous vehicles have seen an increasing interest in recent years, becoming hot topics in both academia and industry. One of the main reasearch areas in this field is related to control systems, in particular planning and decision-making problems. The basic approaches for scheduling high-level maneuver execution modules are based on the concepts of time-to-collision \cite{ttc} and headway control \cite{headway}. In order to add interpretation capabilities to the system, several approaches model the driving decision-making problem as a Partially Observable Markov Decision Process (POMDP, \cite{pomdp}), as in \cite{sdm} for urban scenarios and in \cite{iadd} for intersection handling. A further extension is proposed in \cite{imp} where a \textit{Mixed Observability Markov Decision Process}  (MOMDP) \cite{momdp} is used to model uncertainties in agents intentions. However, since vehicles are assumed to behave in a deterministic way, the aforementioned approaches handle many situations with excessive prudence and would not be able to enter in a busy roundabout. \\
\indent For this reason, the trend of using Deep Learning techniques \cite{dl} for modeling such complex behaviors is growing; in particular \textit{Deep Reinforcement Learning} (DRL) \cite{dlr} algorithms have proved to be efficient even in high-dimensional state spaces and have already been extended to the autonomous driving field, as in \cite{noi} for intersection handling and in \cite{aslc} for lane changes.
However, these works show a major limitation, which is the lack of communication capabilities among vehicles. In fact, those models are trained on synthetic environments in which vehicles movements are based on hard coded rules. A solution to this problem is proposed in \cite{drlad2}, where vehicles inside the simulator were trained through \textit{Imitation Learning} \cite{imitation}; however, this approach is expensive since it requires a huge amount of training data. \\
\indent In our proposed work, this limitation has been overcome training the model in an environment populated by vehicles whose behavior has been learned in a multi-agent fashion as in \cite{mts}. In this way, drivers are able to implicitly communicate through actions and feature a unique, programmable, style of driving, enhancing the realism of the simulation. In order to train agents efficiently in this scenario, a different version of \textit{A3C} \cite{a3c} has been implemented in which the asynchronous agents policies are updated with a lower rate enhancing agents exploration; for this reason it has been called \textit{Delayed A3C} (D-A3C). Desired actions are chosen based on a sequence of images representing what the agent perceives around it. Moreover, our solution permits to set the level of aggressiveness of the artificial driver executing the maneuver; this is essential in those situations in which an excessively cautious behavior or rule-based policies would lead to undefined waits, as in case of insertion in a highly busy roundabout.  \\
\indent As in \cite{sumo} and in \cite{chauffeur}, it has been adopted a simplified synthetic representation of the environment which is easily reproducible by both simulated and real data, so that the system trained offline can be easily validated on a real car equipped with perception systems.
Furthermore, this representation greatly helps in reducing the sample complexity of the problem respect to simulators featuring a realistic graphic such as CARLA  \cite{carla} or GTA-based platforms \cite{GTA}. Our test-bed scenario is the synthetic reconstruction of a real roundabout built with the \textit{Cairo} graphic library \cite{cairo}, shown in Figure~\ref{fig:roundabout}. \\
However, since obstacles detected by the perception systems are not always accurate, our system has been evaluated also in the case of random noise added to the position, size and pose of the vehicles, as well as on the trajectory followed by the agents. \\
The model trained on the single scenario of Figure~\ref{fig:synth_round} has been tested on a different type of roundabout, shown in Figure~\ref{fig:campus_round}, in order to evaluate the generalization capabilities of the system. \\
Finally, tests on real data have been performed using logs recorded with a vehicle equipped with proper sensors.

\section{\uppercase{Background}}
\label{sec:background}
\subsection{Reinforcement Learning}
\begin{figure}
  \centering
  \begin{subfigure}{.49\linewidth}
    \centering
    \includegraphics[width =\linewidth]{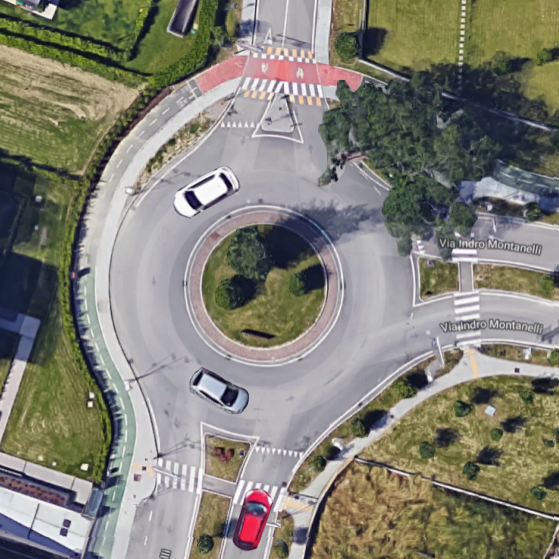}    
    \caption{Real}
    \label{fig:real_world}
  \end{subfigure}
  \begin{subfigure}{.49\linewidth}
    \centering
    \includegraphics[width =\linewidth]{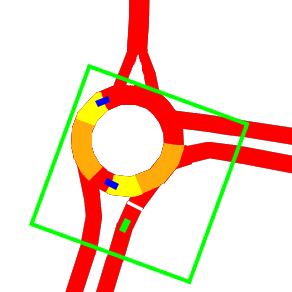}
    \caption{Synthetic}
    \label{fig:synth_round}
  \end{subfigure}
  \vspace{0.5em}
  \caption{Top view of a roundabout (a) and its synthetic representation (b). The green square in (b) highlights the portion of the surrounding perceived by the green vehicle, which is the artificial representation of the red car in (a).}
  \label{fig:roundabout}
\end{figure}

\noindent Reinforcement Learning \cite{suttonbarto} deals with the interaction between an agent and its environment. The actor tries to learn from attempts and errors, receiving a reward signal and observing the state of the environment at every time step. The reward is typically a scalar value and it is related to the sequence of actions taken until that moment. The goal of an agent acting inside an environment at time $t$, is to learn a policy which maximizes the so called \textit{expected return}, which is a function of future rewards; this is generally defined as $R_t = \sum_t^T r_t + \gamma r_{t+1} + \cdots + \gamma^{T-t}r_{T-t}$, where $T$ is the terminal time step and $\gamma$ is a discount factor, used to reduce the importance of future rewards respect to the short-term ones.

\subsection{Multi-agent A3C}
One of the principal difficulties of DRL comes from the strict correlation between consecutive states; initially, the problem was solved by picking up indipendent states from a stored replay buffer \cite{replay}, but this proved to be inefficient in multi-agent scenarios \cite{gupta}. \\
\indent A different approach is taken in A3C \cite{a3c}, where several copies of the agent take actions in parallel, so that each one experiences states of the environment which are independent from those of the others, enhancing the stability of the learning process. Agents send their updates and amend their local copy of the network every \textit{n}-step frames. \\
\indent Multi-agent A3C \cite{mts} follows the same principle, but allows some of the agents to share the same instance of the environment, inducing them to learn how to interact in order to commonly achieve their goals. Thus, an implicit agent-to-agent negotiation can gradually arise, since actions taken from an agent will affect the state of others and vice-versa.

\subsection{A2C}
A2C \cite{a2c} is the synchronous variant of A3C in which agents compute and send their updates at fixed time intervals. This solution is more time-efficient because it permits the computation of updates of all agents in a single pass exploiting GPU computing. However, since all agents hold the same policy, the probability of converging on a local minimum of the loss function may increase, altough it has not yet proven empirically \cite{a2c}.

\section{\uppercase{Decision-Making Module}}
\label{sec:decision_making_module}
\subsection{D-A3C Implementation}
\begin{figure}[b]
  \centering
  \begin{subfigure}{.33\linewidth}
    \centering
    \includegraphics[width =\linewidth]{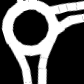}    
    \caption{Navigable space}
    \label{fig:nav_space}
  \end{subfigure}
  \hspace{0.5cm}
  \begin{subfigure}{.33\linewidth}
    \centering
    \includegraphics[width =\linewidth]{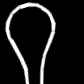}
    \caption{Path}
    \label{fig:path}
  \end{subfigure}
  \begin{subfigure}{.33\linewidth}
    \centering
    \vspace{1em}
    \includegraphics[width =\linewidth]{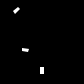}    
    \caption{Obstacles}
    \label{fig:obstacles}
  \end{subfigure}
  \hspace{0.5cm}
  \begin{subfigure}{.33\linewidth}
    \centering
    \vspace{1em}
    \includegraphics[width =\linewidth]{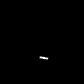}
    \caption{Stop line}
    \label{fig:stop_line}
  \end{subfigure}
  \vspace{0.5em}
  \caption{Semantic layers of the input space of the agent.}
  \label{fig:ab}
\end{figure}
\noindent Our module is trained by a reinforcement learning algorithm which we will refer to as \textit{Delayed A3C} (D-A3C), where the goal of the agent, called \textit{active}, is a safe insertion in a roundabout populated by other vehicles, the \textit{passives}, already trained in a multi-agent fashion in order to be endowed with interaction capabilities \cite{mts}. The actions performed by the \textit{active} for entering the roundabout are controlled by the output of a neural network whose architecture is similar to the one adopted in \cite{mts}. 

Our implementation differs from the original A3C in the way the asynchronous learners update the global neural network collecting all the actors' contributions. Indeed, our learners exchange the computed updates with the global network only at the end of their episode, keeping the same policy for the whole episode execution, while classic A3C does it at fixed and shorter time intervals. This reduces the synchronization burden of the algorithm, since the number of parameter exchanges diminishes. Moreover, in Section~\ref{sec:algo_comparison} we demonstrate that D-A3C leads to better results than classical A3C in the analyzed task. We did not carry further tests for evaluating the performances of the two algorithms in other tasks, since it is not the scope of this work; however, this comparison could be the subject of future studies.

The environment in which an agent is learning can be different from that of another agent, permitting to train the policy in a range of different scenarios simultaneously. Indeed, in our experiments we teach the agents how to enter in a three-entry-roundabout from all the entries simultaneously; nonetheless, in order to achieve a sufficient amount of agents for the learning process to be stable, we let multiple copies of the agents to learn from every entry in indipendent copies of the roundabout. \\
Multi-environment architectures should increase the model generalization capabilities: this is tested in Section~\ref{sec:unknown_round} where the performance of our system is evaluated on an unseen roundabout.

\subsection{Input Space}
The input space of the system is composed by two different types of streams: a visual and a non-visual sensory channel. The visual input is a sequence of four images having size 84x84x4, that is a mapping of the 50x50 meters of the vehicle's sourrounding. These images represent 4 semantic layers consisting in the navigable space (Figure~\ref{fig:nav_space}) in which the agent can drive, the path (Figure~\ref{fig:path}) that the agent should follow, the obstacles (Figure~\ref{fig:obstacles}) around the agent including itself and the stop line (Figure~\ref{fig:stop_line}) that is the position where the agent should stop if the entry cannot be made safely. \\
On the other hand, the non-visual sensory channel is composed by 4 entities: the first one is the \textit{agent speed}, that is the absolute value of the current speed of the agent; the second one is the \textit{target speed}, that represents the maximum speed that the actor should reach and maintain if there is no traffic and enough visibility; the third one is the \textit{aggressiveness}, namely the degree of impetus in the maneuver execution and the last one represents the last action performed by the agent.
\label{sec_input_space}

\subsection{Output}
The output of the maneuver planning system is a prediction over the following states:
\begin{itemize}
\item \textbf{Permitted}: the agent perceives the entry area of the roundabout as free and entering would not create any dangerous situation. This state sets the acceleration $a$ of the \textit{active} vehicle to a fixed comfort value $a_{max}$ unless the \textit{target speed} is reached.
\item \textbf{Not Permitted}: the agent predicts the entry area of the roundabout as busy and entering would be dangerous. This state produces a deceleration computed as $\min(d_{max}, d_{stop\_line})$, where $d_{max}$ is the maximum deceleration permitted following comfort constraints, and $d_{stop\_line}$ is the deceleration needed to arrest the \textit{active} vehicle at the stop line. If the agent has already overcome the stop line, this state causes a brake of a $d_{max}$ value.
\item \textbf{Caution}: the roundabout is perceived as not completely free by the \textit{active} agent and the vehicle should approach it with prudence, either to improve the view for a safe entering or to observe if an oncoming \textit{passive} vehicle is willing to let it enter the roundabout; the maximum speed permitted to the agent is $\frac{1}{2}$ \textit{target speed} and $a$ can assume one of the following values:
  \begin{equation}
    a=
    \begin{cases}
      \frac{a_{max}}{2}, & \textbf{if}\ agent\ speed < \frac{target\ speed}{2} \\
      \frac{d_{max}}{2}, & \textbf{if}\ agent\ speed > \frac{target\ speed}{2} + h \\
      0, & \textit{otherwise}
    \end{cases}
  \end{equation}

where $h$ is a costant set to 0.5.

\end{itemize}

\subsection{Reward}
\label{reward}
The reward $r_t$ is composed by the following terms:
\begin{equation} \label{eq:rt}
r_t = r_{danger} + r_{terminal} + r_{indecision} + r_{speed}
\end{equation}

\noindent $r_{danger}$ is a penalization given to the \textit{active} agent when it performs dangerous maneuvers and it is defined as:
\begin{equation}
r_{danger} = -w_{d_s} \cdot \alpha \cdot d_s - w_{c_f} \cdot \alpha \cdot c_f
\end{equation}
in which,
\begin{itemize}
\item $d_s$ is a binary variable which is set to 1 when the \textit{active} vehicle violates the safety distance from the \textit{passive} one in front; this distance is equal to the space traveled from the \textit{active} agent in one second, as shown from the yellow region in Figure~\ref{fig:synth_round}. When the safety distance is maintained the value of $d_s$ is 0;
\item $c_f$ is a binary variable and it is set to 1 when the \textit{active} agent cuts in front of a \textit{passive} vehicle already inserted in the roundabout; this region is equal to three times the distance traveled from the \textit{passive} vehicle in one second. This is shown from the orange region in Figure~\ref{fig:synth_round}. If the learning actor does not break this rule the value of $c_f$ is 0.
\item $\alpha$ depends on the aggressiveness level of the \textit{active} agent and it is defined as ${\alpha = (1 - aggressiveness)}$. During the training phase, $aggressiveness$ assumes a random value from 0 to 1 kept fixed for the whole episode. Higher values of $aggressiveness$ should encourage the actor to  increase the impatience; consequently, dangerous actions will be less penalized.
In the test phase we fix the $aggressiveness$ value in order to perform comparisons among agents with different values of this parameter, as shown in Section~\ref{aggressiveness_test}.
\item $w_{d_s}$ and $w_{c_f}$ are weights set to 0.002 and 0.005 respectively.
\end{itemize}

\noindent $r_{terminal}$ depends on how the episode ends. In order to avoid an excessively conservative behavior of the \textit{active} agent, it is imposed a maximum available time for the actor to reach its target. The possible values $r_{terminal}$ can assume are:
\begin{itemize}
\item +1: if the \textit{active} agent ends the episode safely, reaching its goal;
\item $-\beta - \gamma \cdot \alpha$: if the \textit{active} actor does not finish the episode because of a crash with another agent. $\beta$ is a costant set to 0.2, while $\gamma$ is the weight of $\alpha$ set to 1.8. Hence, when a crash occurs, we modulate $r_{terminal}$ based on the $aggressiveness$, for the same reason explained for $r_{danger}$.
\item -1: if the time available to finish the episode expires.
\end{itemize}

\noindent $r_{indecision}$ is a negative reward in order to provide a realistic and smooth behavior to the agent, avoiding frequent changes of conflicting actions. It depends on the last two states of the system: we penalize the agent when the state passes from \textit{Permitted} to one of the others. Calling $L1$ and $L2$ the last and the second to last outputs respectively, we can resume this reward with the following equation:
  \begin{equation}
    r_{indecision}=
    \begin{cases}
      - 0.05, \ \textbf{  if} & L2=\textit{Permitted}  \\& \text{and } L1=\textit{Caution} \\
      - 0.15, \ \textbf{  if} &  L2=\textit{Permitted} \\& \text{and } L1=\textit{Not Permitted} \\
      0, & \textit{otherwise}.
    \end{cases}
  \end{equation}

\noindent $r_{speed}$ is a positive reward which encourages the \textit{active} vehicle to increase the speed. It is defined as:
\begin{equation}
r_{speed} = \psi \cdot \frac{current\ speed}{target\ speed}
\end{equation}
in which $\psi$ is a constant set to 0.0045 and the \textit{target speed} at the denominator acts as a normalizing factor. \\
This reward shaping is essential to ensure that the agents learn the basic rules of the road like the right of way and the safety distance.

The following link (\url{https://youtu.be/qVwRCad5K9c}) shows how the \textit{active} agent performs the entering in the roundabout.

\begin{table*}[b] \centering
\caption{Comparison between D-A3C model and rule-based approach.}
\label{table1}
\begin{tabular}{|c|c|c|c|c|c|}
\hline
 & 
\multicolumn{4}{c|}{Rule-based} & 
\multirow{2}{*}{D-A3C} \\
&
\multicolumn{1}{c}{25m} &
\multicolumn{1}{c}{20m} &
\multicolumn{1}{c}{15m} &
\multicolumn{1}{c|}{10m} &
\multicolumn{1}{c|}{} \\
\hline
Reaches \% & 0.456 & 0.732 & 0.831 & 0.783 & 0.989 \\
\hline
Crashes \% & 0.0 & 0.002 & 0.012 & 0.100 & 0.011 \\
\hline
Time-overs \% & 0.544 & 0.266 & 0.157 & 0.117 & 0.0 \\
\hline
\end{tabular}
\end{table*}

\section{\uppercase{Experiments}}
\label{sec:experiments}
\subsection{Algorithms Comparison}
\label{sec:algo_comparison}
\begin{figure}[h]
\centering
\includegraphics[width=0.8\linewidth]{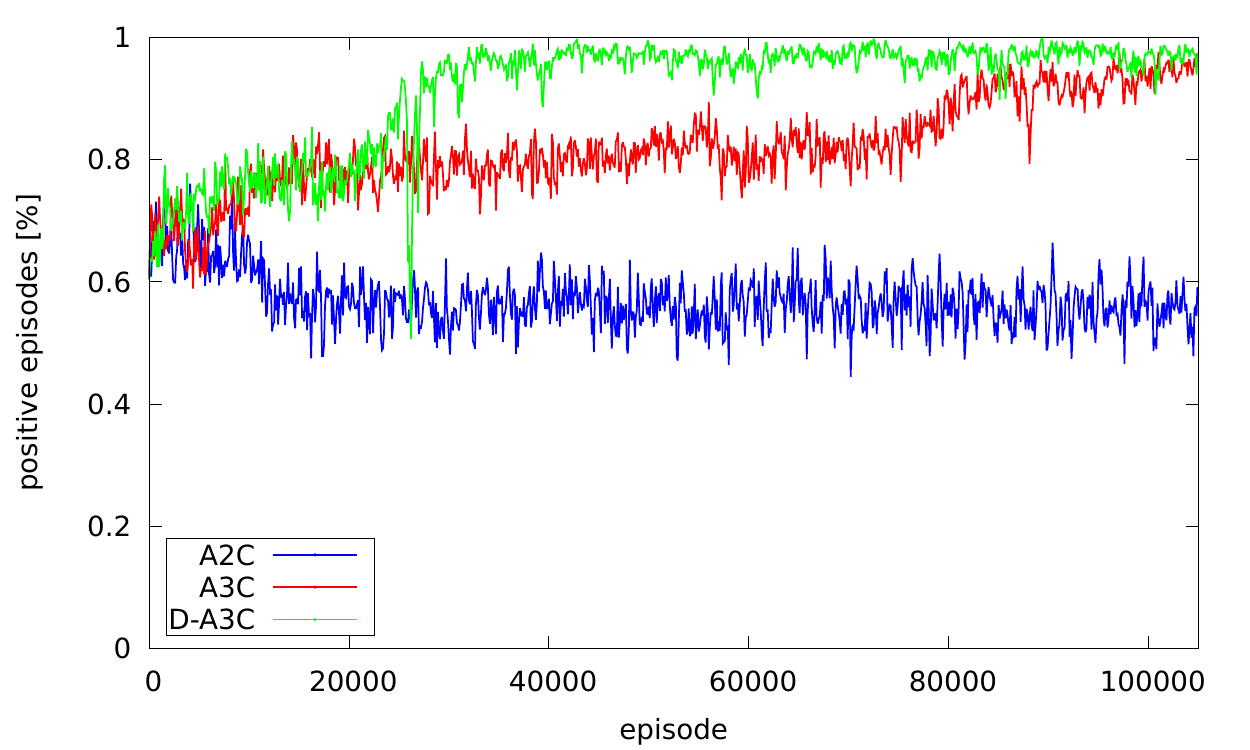}
\caption{Moving average of the positive episodes ratio using D-A3C (green), A3C (red) and A2C (blue).}
\label{fig:algo_comparison}
\end{figure}

We compared the A3C and A2C algorithms with our D-A3C implementation in order to test if our implementation improves the learning performances. The curves of Figure~\ref{fig:algo_comparison} show that A3C needs more episodes than our method for learning successfully the task. Instead, A2C converges on a suboptimal solution, consisting on always outputting the \textit{Permitted} state, letting the agent entering the roundabout independently on the occupancy of the road. The \textit{aggressiveness} used during the training phase is chosen randomly ([0, 1]) and kept fixed during each episode both for \textit{passive} and for \textit{active} vehicles, while the maximum number of \textit{passive} vehicles populating the roundabout (Figure~\ref{fig:synth_round}) simultaneously is set to 8.

\subsection{Comparison with a Rule-based Approach}
\label{sec:rule_based}

The metrics used to evaluate the performances are \textit{Reaches}, \textit{Crashes} and \textit{Time-overs} corresponding to the percentages of episodes ended successfully, with a crash and due to the depletion of the available time respectively. Every test is composed by three experiments (each one composed by 3000 episodes) using three different traffic conditions: \textit{low}, \textit{medium} and \textit{high} which correspond to a maximum number of \textit{passive} agents populating the roundabout to 4, 6 and 8 respectively. The results in Table~\ref{table1} represent the average percentages of the three experiments. We compared the results obtained by D-A3C model on the training roundabout (Figure~\ref{fig:synth_round}) with those achieved by a simple rule-based approach. In particular, we set four tresholds (25, 20, 15 and 10 meters) corresponding to the minimum distances required from a \textit{passive} vehicle to the \textit{active} one for starting the entering maneuver. Even if the percentages of crashes are rather low, the results in Table~\ref{table1} show that a rule-based approach could lead to long waits since its lack of negotiation and interaction capabilities brings the agent to perform the entry only when the roundabout is completely free.

\begin{figure}[b]
\centering
\includegraphics[scale=0.50]{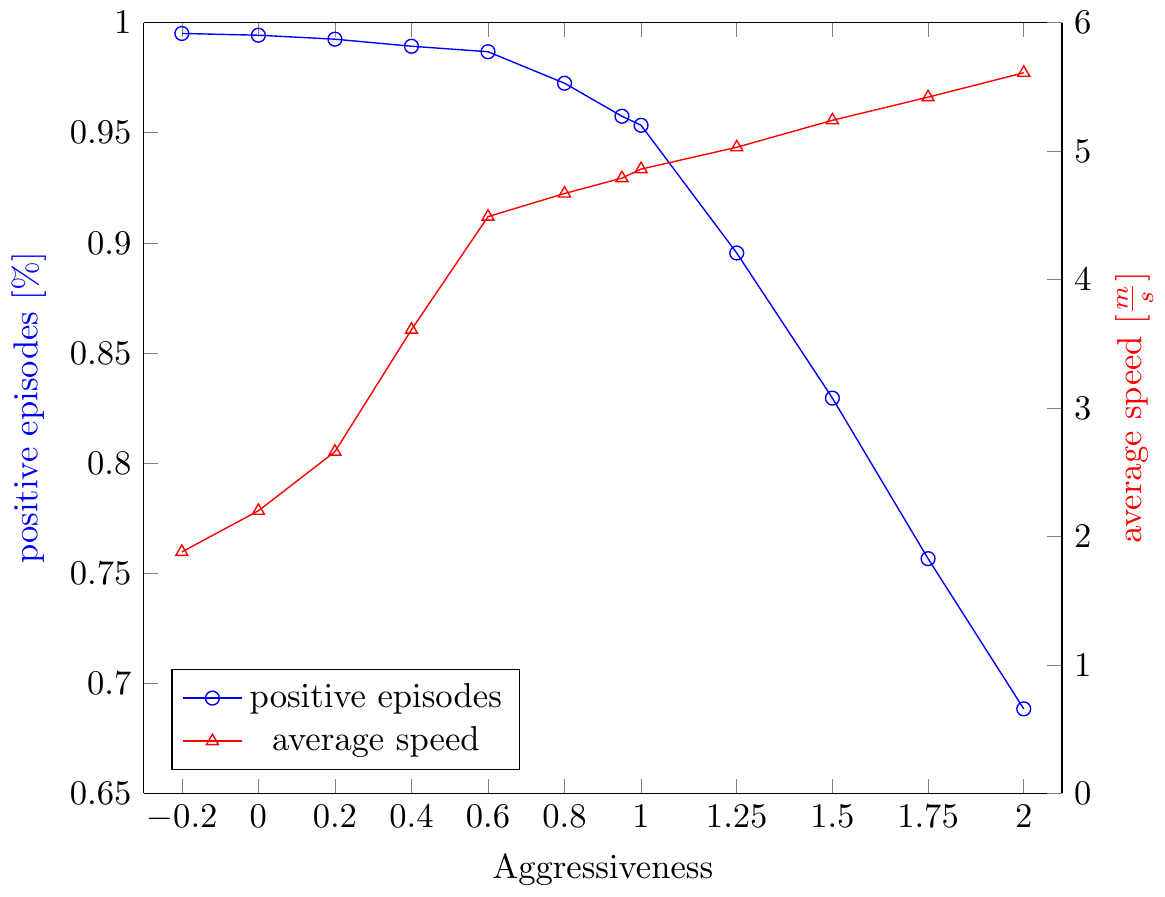}
\vspace{0.5em}
\caption{Values of average speed and positive episodes ratio depending on the aggressiveness level of the active agent.}
\label{fig:aggressiveness}
\end{figure}

\subsection{Aggressiveness Tests}
\label{aggressiveness_test}
As explained in Section~\ref{sec_input_space}, the aim of the \textit{aggressiveness} input is to give the possibility of modulating the agent behavior depending on the traffic conditions. 
This is achieved by shaping the rewards accordingly to this input during the training phase as explained in Section~\ref{reward}, and exposing the agent to different traffic conditions. In order to prove the efficacy of the \textit{aggressiveness} tuning, we tested the \textit{D-A3C} model on a busy roundabout varying the aggressiveness level from low to high, highlighting the full spectrum of behaviors.
We calculated the \textit{average speed} of the \textit{active} vehicle and the ratio of the episodes which ended successfully. As can be noted from Figure~\ref{fig:aggressiveness}, the aggressiveness input acts a crucial role in determining the output of the module: higher values of this input rise the impatience of the \textit{active} vehicle which tends to increase the risks taken. This produces an increment of crashes with a consequent decrease of the \textit{positive episodes} ratio, but also an increase of the \textit{average speed} value. In real-world tests, higher values of \textit{aggressiveness} can be useful in deadlock situations (for example high traffic condition), flanking the module with safety systems in order to avoid collisions. \\
Moreover, it is interesting to notice that the behavior of the system is coherent also for those \textit{aggressiveness} values outside the range used during the training phase ($[0, 1]$).

\begin{table}[!b] \centering
\caption{Results on the unknown roundabout.}
\label{table2}
\begin{tabular}{|c|c|c|c|}
\hline
 & {D-A3C} & {Random} & {Permitted}\\
\hline
Reaches \% &  0.910 & 0.684 & 0.676 \\
\hline
Crashes \% & 0.085 & 0.270 & 0.324  \\
\hline
Time-overs \% & 0.003 & 0.046 & 0.0 \\
\hline
\end{tabular}
\end{table}

\subsection{Performances on Unknown Roundabouts}
\label{sec:unknown_round}

\begin{figure}[!b]
  \centering
  \begin{subfigure}{.4\linewidth}
    \centering
    \includegraphics[width =\linewidth]{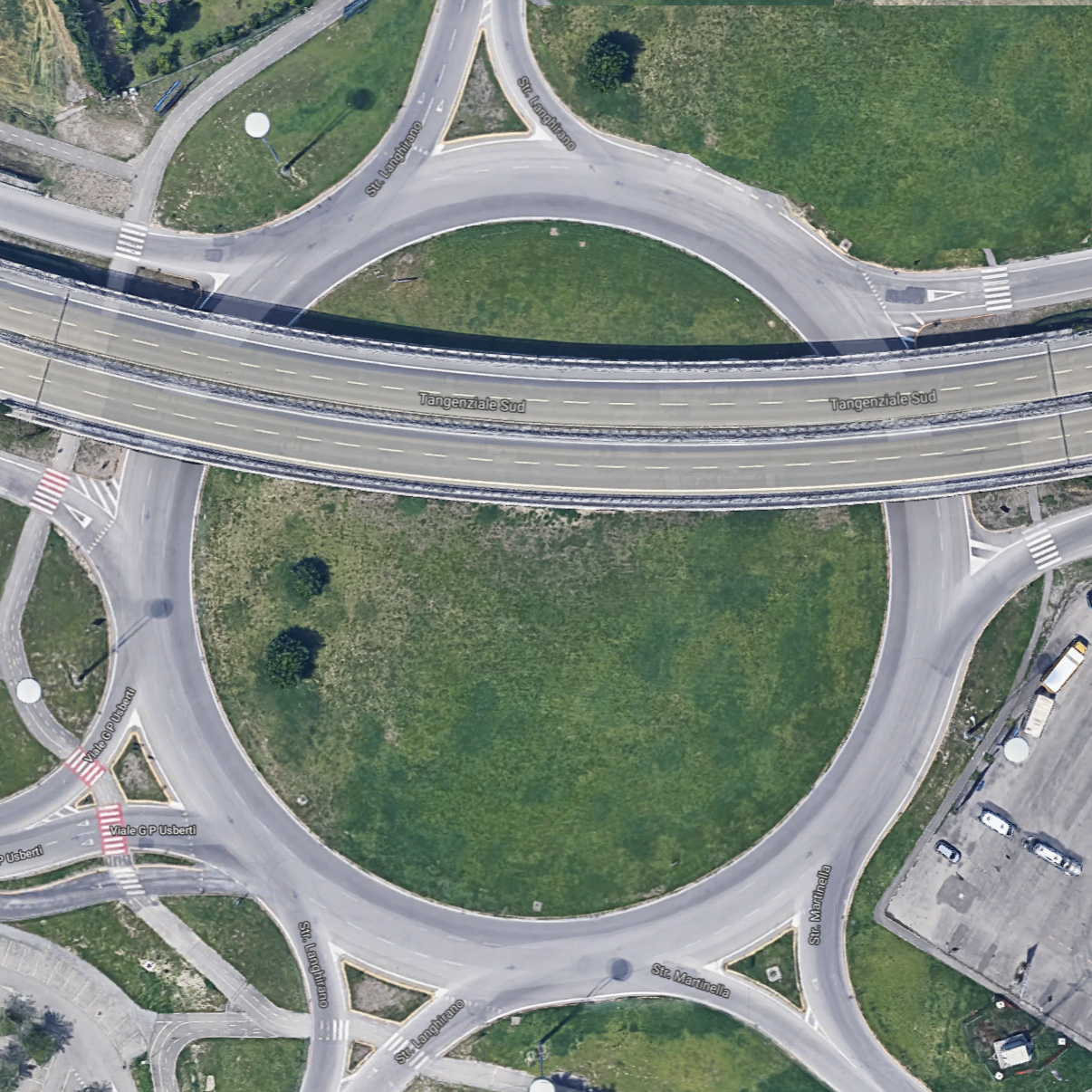}    
    \caption{Real}
    \label{fig:real_world_campus}
  \end{subfigure}
  \hspace{0.5cm}
  \begin{subfigure}{.4\linewidth}
    \centering
    \includegraphics[width =\linewidth]{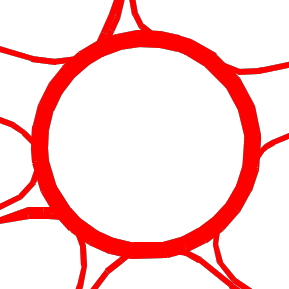}
    \caption{Synthetic}
    \label{fig:synth_round_campus}
  \end{subfigure}
  \vspace{0.5em}
  \caption{Top view of the real (a) and its synthetic representation (b) of the roundabout which was not seen by the agent during training.}
  \label{fig:campus_round}
\end{figure}

We tested the system on a different type of roundabout from the one used in the training phase; the new roundabout, shown in Figure~\ref{fig:campus_round}, features a different shape and number of entries.
We compared the results achieved by the model with two different baselines: the first one obtained fixing the output of the module to the \textit{Permitted} state independently on the occupancy of the road and the second one obtained with random actions. Due to a larger area involved in this test, \textit{low}, \textit{medium} and \textit{high} traffic conditions correspond to a maximum number of passive agents inside the roundabout to 10, 15 and 20 respectively. The results in Table~\ref{table2} show that the system features some generalization capabilities; however, considering the results achieved by \textit{D-A3C} model on the training environment (Table~\ref{table1}), we can observe that the diversity of the training set environments has to be increased in order to improve the performances of the system in unseen roundabouts.

\begin{table}[!b] \centering
\caption{Results on the noised environment.}
\label{table3}
\begin{center}
\begin{tabular}{|c|c|c|}
\hline
 & {D-A3C} & {Noised D-A3C} \rule{0pt}{2ex} \\
\hline
{Reaches \%} &  0.899 & 0.967 \rule{0pt}{2ex} \\
\hline
{Crashes \%} & 0.043  & 0.021 \rule{0pt}{2ex} \\
\hline
{Time-overs \%} & 0.058  & 0.012 \rule{0pt}{2ex} \\
\hline
\end{tabular}
\end{center}
\end{table}

\begin{figure}[!b]
\centering
\includegraphics[width=0.15\textwidth]{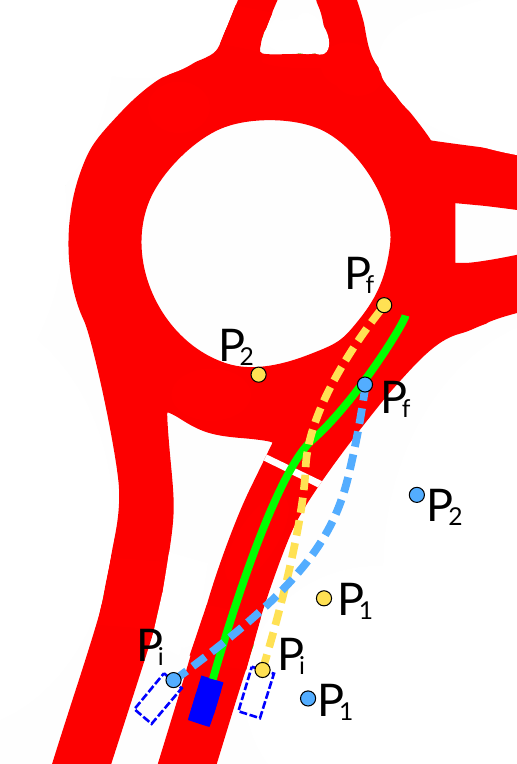}
\vspace{0.5em}
\caption{Example of Bézier curves: the green line (solid) represents the original path, while the light blue and the yellow lines (dotted) represent two possible Bézier curves.}
\label{fig:bezier_curves}
\end{figure}

\subsection{Perception-noise Injection}
\begin{figure*}[h]
\centering
\includegraphics[width=0.81\textwidth]{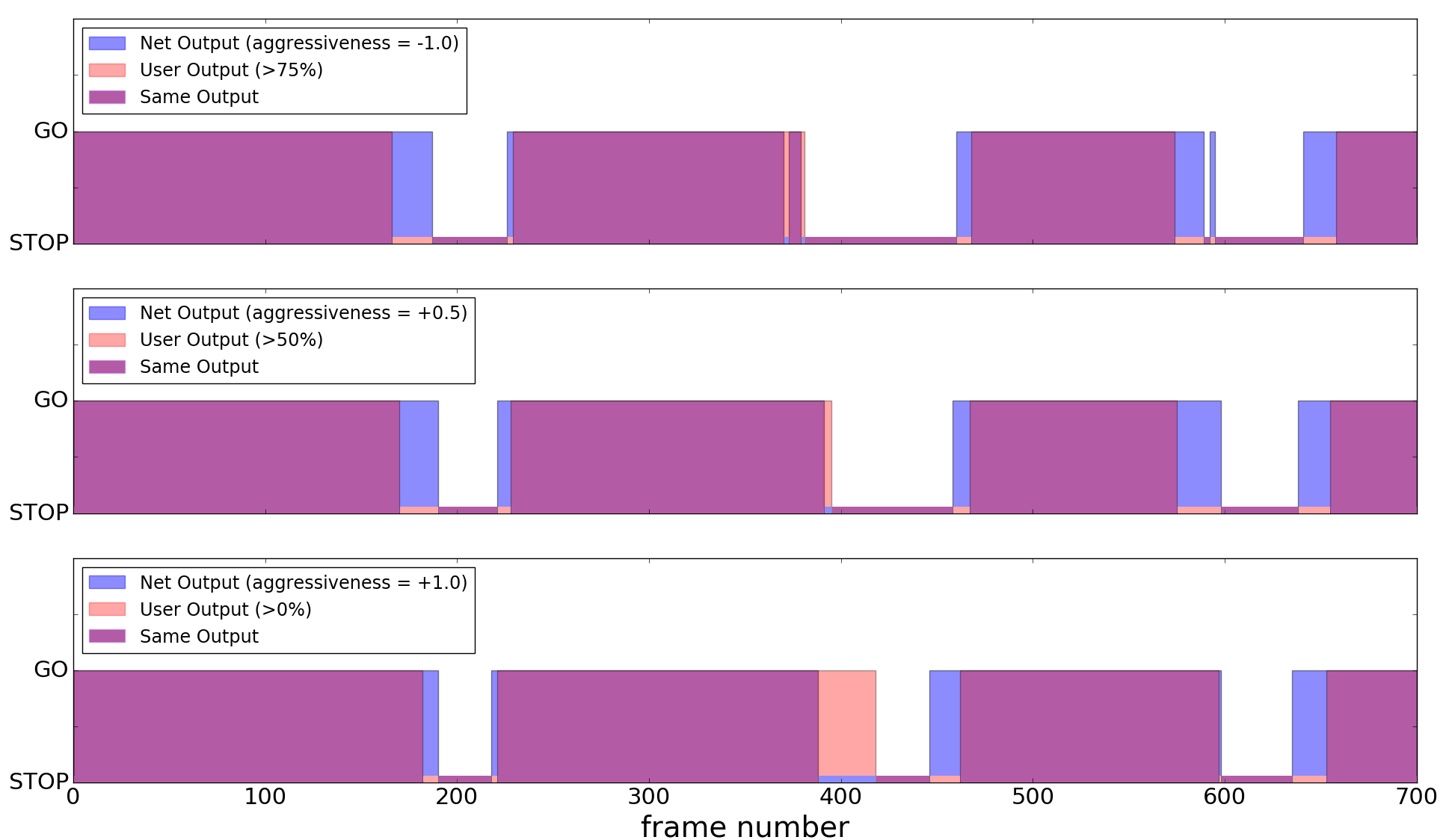}
\vspace{0.0em}
\caption{Comparison between the output of D-A3C module and those of the users for the first 700 frames (out of 2000). The blue and the red areas correspond to the decisions of our system and the users choices respectively, while the violet ones represent the frames in which the users and the net perform the same actions.}
\label{fig:net_user_comparisons}
\end{figure*}

We introduced two types of noise in order to reduce the gap between synthetic and real data. We added gaussian noise in the position, size and pose estimation of \textit{passive} agents to simulate those errors of the systems on-board the real vehicle. Then, we perturbed the path of \textit{active} agents with Cubic Bézier curves computed by the De Casteljau algorithm \cite{de_casteljau}, in order to avoid following the same route as happens in the real world. This noise is also useful to make the system more robust to localization errors that may occur during tests on a self-driving vehicle.
As shown in Figure~\ref{fig:bezier_curves}, for each episode we randomly chose the initial and the final points, called $P_i$ and $P_f$ respectively; the only constraints are that
\begin {enumerate*} [label=\itshape\alph*\upshape)]
\item $P_i$ ranges between the first point of the original path (the green line in Figure~\ref{fig:bezier_curves}) and the stop line; \item $P_f$ ranges between the stop line and the last point of the path. 
\end {enumerate*} 
Finally, we calculated two anchors $P_1$ and $P_2$ choosing two random points along the path and perturbating their coordinates $(x, y)$ with gaussian noise. \\
Starting from the \textit{D-A3C} model, we used \textit{Curriculum Learning} \cite{cl} to train the system in the noisy environment obtaining a new model which we call \textit{Noised D-A3C}. We evaluated the two models in the noised environment, performing tests as in Section~\ref{sec:rule_based}; the results in Table~\ref{table3} show that the \textit{Noised D-A3C} model becomes more robust to localization and perception errors. However, further tests on how to achieve better generalization on real data will be performed in future works.

\subsection{Test on Real Data}
We evaluated our module with real data recording both around 2000 perception frames and video streams of the roundabout of Figure~\ref{fig:real_world} with a car equipped with a stereo camera and a GPS. We projected the recorded traffic into our synthetic environment and performed a test with three different aggressiveness levels of the net ($-1.0, +0.5, +1.0$). These values represent different driving styles, from the most cautious to the most impetuous; since in Section~\ref{aggressiveness_test} we noticed that the behavior of the system is coherent also for values outside the range used during the training, we chose $-1.0$ to accentuate the cautious behavior. \\
The frames have been recorded with the car stopped at the stop line and therefore also the agent in our synthetic environment took its decisions from the same point. However, in this way it is possible to evaluate the single-shot insertion but not the full capabilities of the system. In order to compare these results with human decisions, we developed a simple interface in which users, watching the real recorded sequences, have to choose when to enter in the roundabout and when to stay stop. However, since the output of our system is a prediction of three possible states (\textit{Permitted}, \textit{Not Permitted} and \textit{Caution}), we modeled the \textit{Caution} state as \textit{Not Permitted} to make a correct comparison between users and \textit{D-A3C} model actions.\\
We stored the decisions of 10 users and we set up a counter for each frame, representing the number of users that would perform the entry in the roundabout at that time, such that its value ranges from $0$ to $10$. We created three different artificial user profiles based on these counter values: the first one in which the entry is performed if at least the 75\% of the users would enter in the roundabout, the second one with this ratio equal to 50\% and the last one in which at least one user ($>$0\%) would enter in the roundabout. These percentages represent three different human driving styles such that we could compare them with the different aggressiveness levels of the net explained previously. Figure~\ref{fig:net_user_comparisons} illustrates the comparison between the actions of our module (blue) and those of the users (red) in each of the three profiles. Moreover, Table~\ref{table4} shows the average match percentages over the three video sequences between the first, the second and the third user profile with the results obtained setting the aggressiveness level of our system to $-1.0$ (\textit{Comparison \#1}), $+0.5$ (\textit{Comparison \#2}) and $+1.0$ (\textit{Comparison \#3}) respectively. Since the match percentages between different single user decisions range from 80\% to 95\%, we can observe that the results achieved in Table~\ref{table4} represent a good match between our module output and human decisions.

\begin{table}[!b] \centering
\caption{Average match percentages between user profiles and \textit{D-A3C} model actions over the three video sequences.}
\label{table4}
\begin{center}
\begin{tabular}{|c|c|}
\hline
{ Comparison \#1\%} & {81.288} \rule{0pt}{2ex} \\
\hline
{ Comparison \#2\%} & {84.389} \rule{0pt}{2ex} \\
\hline
{ Comparison \#3\%} & {84.515} \rule{0pt}{2ex} \\
\hline
\end{tabular}
\end{center}
\end{table}

\section{\uppercase{Conclusion}}
\noindent In this paper we presented a decision-making module able to control autonomous vehicles during roundabout insertions. The system was trained inside a synthetic representation of a real roundabout with a novel implementation of A3C which we called \textit{Delayed A3C}; this representation was chosen so that it can be easily reproduced with both simulated and real data. The developed module permits to execute the maneuver interpreting the intention of the other drivers and implicitly negotiating with them, since their simulated behavior was trained in a cooperative multi-agent fashion. \\
We proved that D-A3C is able to achieve better learning performances compared to A3C and A2C by increasing the exploration in the agent policies; moreover, we demonstrated that negotiation and interaction capabilities are essential in this scenario since a rule-based approach leads to superfluous waits. \\
It also emerged that the decision-making module features light generalization capabilities both for unseen scenarios and for real data, tested by introducing noise in the obstacles perception and in the trajectory of agents. However, these capabilities should be enforced in future works for making the system usable both in real-world and unseen scenarios.\\
Finally, we tested our module on real video sequences to compare the output of our module with the actions of 10 users and we observed that our system has a good match with human decisions.

\bibliographystyle{apalike}
{\small
\bibliography{example}}

\end{document}